\documentclass[conference]{IEEEtran}
\IEEEoverridecommandlockouts
\usepackage{cite}
\usepackage{amsmath,amssymb,amsfonts}
\usepackage{algorithmic}
\usepackage{graphicx}
\usepackage{textcomp}
\usepackage{xcolor}
\usepackage{subcaption}

\usepackage{times}
\usepackage{parskip}
\usepackage{subfig}
\usepackage{psfrag}
\usepackage{wrapfig}
\usepackage[latin1]{inputenc}
\usepackage{amssymb}
\usepackage{epsfig}
\usepackage{graphicx,times} % Add all your packages here
\usepackage{amsfonts,amsmath,amssymb,array}
\usepackage{tabularx}   % Erweiterte Tabellen Optionen
\usepackage{colortbl}
\usepackage{longtable}
\usepackage{booktabs}
\usepackage{url}
\usepackage{icomma}
\usepackage{pstricks}
\usepackage{longtable}
\usepackage{multirow}
\usepackage{color}
\usepackage{graphicx}
\usepackage{cite}
\usepackage{picinpar}
\usepackage{amsmath}
\usepackage{amssymb}
\usepackage{amsthm}
\usepackage{mathtools}
\usepackage{url}
\usepackage{flushend}
\usepackage[latin1]{inputenc}
\usepackage{colortbl}
\usepackage{soul}
\usepackage{multirow}
\usepackage{pifont}
\usepackage{color}
\usepackage{alltt}
\usepackage{breakurl}
\usepackage{epstopdf}
\usepackage{pbox}
\usepackage{bbm}
\usepackage{algorithm2e}
\RestyleAlgo{ruled}

\def\BibTeX{{\rm B\kern-.05em{\sc i\kern-.025em b}\kern-.08em
    T\kern-.1667em\lower.7ex\hbox{E}\kern-.125emX}}

\begin{document}

\title{Distributed Optimization of Modular Production Systems using Model-based Reinforcement Learning with Inverse Models\\
}

\author{
\IEEEauthorblockN{Andreas Schwung\IEEEauthorrefmark{1}, Steve Yuwono \IEEEauthorrefmark{1}, Sofiene Lassoued \IEEEauthorrefmark{1}, Dorothea Schwung\IEEEauthorrefmark{2}}% <-this % stops a space
%\thanks{*This work was not supported by any organization}% <-this % stops a space
\IEEEauthorblockA{\IEEEauthorrefmark{1}South Westphalia University of Applied Sciences, Soest 59494, Germany
    \\\{schwung.andreas, yuwono.steve, lassoued.sofiene \}@fh-swf.de} \IEEEauthorblockA{\IEEEauthorrefmark{2} Hochschule D{\"u}sseldorf University of Applied Sciences, D{\"u}sseldorf 40476, Germany
\\ dorothea.schwung@hs-duesseldorf.de}
}

\maketitle

\begin{abstract}
This paper presents a novel approach for data-driven self-learning control of highly flexible, modular manufacturing systems. Specifically, we employ a novel framework for model-based reinforcement learning which introduces approximate inverse process models within the training of reinforcement policies. This approach disentangles the learning of actuation dynamics and the dynamics in state space, resulting in RL-based training solely within the task space. We propose a lightweight feedforward architecture for approximate inverse models and integrate them within the policy network of standard RL algorithms. We apply the approach to a laboratory modular production testbed with heterogeneous production modules. The results underline the efficiency improvements for modular manufacturing units in terms of both performance and training speed, particularly for off-policy algorithms.
\end{abstract}


\section{Introduction}

Complex demands on modern manufacturing processes, specifically the requirements for small lot size production of individualized products, require both highly digitalized production units as well as self-learning and self-reconfiguration capabilities. Such fast-adapting systems offer plug-and-play functions~\cite{Schleipen2015} such that near-zero reconfiguration and fast adjustment to different production goals is achievable. 

A fundamental prerequisite for the realization of such production systems is a consistently modular system architecture encompassing both the hardware and software levels. While this modularity significantly enhances the flexibility and scalability of the system, it also increases the complexity of its control. Consequently, decentralized and distributed control and communication concepts are required, which are commonly based on the paradigm of Multi-Agent Systems (MAS)~\cite{10.1561/2600000019}. Furthermore, dynamic production environments, continuously evolving requirements, and the demand for resource-efficient and highly productive manufacturing necessitate the integration of self-learning capabilities into the control architecture. To realize such self-learning functionality, various machine learning approaches can be employed, which are generally categorized into supervised learning, unsupervised learning, and reinforcement learning (RL), see~\cite{Wang2018,Wuest2016} for recent overviews.

Among these techniques, RL has emerged as a particularly promising approach for equipping dynamic systems with self-learning capabilities. In RL, an autonomous agent learns to select actions through repeated interactions with its environment, with the objective of maximizing the cumulative reward obtained over time~\cite{Sutton1998}. Building upon the traditional single-agent framework, reinforcement learning has been extended to multi-agent settings, resulting in the field of Multi-Agent Reinforcement Learning (MARL)~\cite{Busoniu2008}. Since flexible manufacturing systems are typically characterized by distributed decision-making and decentralized control structures, the MARL paradigm provides a natural and well-suited framework for enabling autonomous learning and coordination in such environments.

Despite the significant progress achieved in RL, its adoption in industrial applications remains limited. This is primarily due to inherent challenges that restrict its practical deployment. One major limitation is the extensive training effort required by most RL algorithms~\cite{Schwung2018a}. Moreover, state-of-the-art deep RL methods require large amounts of training data before satisfactory performance can be achieved. In industrial settings, acquiring such comprehensive datasets from real production processes is impractical due to limited operating time, associated costs, and the risk of production interruptions.

One approach to reduce both data and training times is model-based RL (MBRL) in which a model of the system is first learned in a supervised manner using a comparably small dataset from random interactions with the environment. The RL agent is then trained using the model instead of the real environment, thereby significantly reducing the environment interactions. While MBRL is widely applied, the potential of inverting the learned system models to predict appropriate actions given a desired state trajectory has rarely been explored. 

In this paper, we bridge this gap by proposing the integration of (approximately) inverse models of the forward system model into policy networks. This allows the use of trained relations between states and actions in both directions, resulting in a partly pretrained policy network. Specifically, this allows us to restrict the learning behavior of the agent to learn only in the task space, meaning the RL agent mainly trains desired state trajectories instead of the more complex to train action trajectories.  

The contributions of the paper can be summarized as follows:
\begin{itemize}
    \item We propose a novel approach to MBRL which employs inverse models within policy networks, thereby disentangling the actuation dynamics from task space learning. 
    \item We propose a lightweight network architecture for (approximate) model inversion based on feedforward neural networks. 
    \item We present the application of the novel approach on a modular production system with very encouraging results in terms of both performance and learning speed.
\end{itemize}

The paper is structured as follows: Section~\ref{sec:rel} discusses the related work. Section~\ref{sec:DMBRL} provides the problem statement and presents the proposed framework. Section~\ref{sec:results} provides results and comparisons while Section~\ref{sec:concl} concludes the paper.

\section{Related Work}\label{sec:rel}

We review literature on both distributed learning in manufacturing systems and subsequently MARL and MBRL approaches.

\subsection{Distributed Learning in Manufacturing Systems}\label{sec:rel2}

Distributed manufacturing systems (DMS) are production processes coordinated in multiple decentralized locations~\cite{BORTOLINI201893}. Unlike traditional systems that are limited to a single location, DMS distribute decision-making and production processes among multiple nodes. Conventional approaches for distributed optimization and control include distributed optimal control~\cite{10608415}, distributed Model Predictive Control~\cite{9640463}, consensus-based distributed control~\cite{10948010}, and Fuzzy Control~\cite{11084911}. However, these approaches typically lack self-learning capabilities and hence, cannot be applied to fast-changing manufacturing systems with varying behavior. Specifically, as they typically rely on fixed models, they are inherently exposed to differences between model and real world.

Recently, self-learning systems like MARL and game theory-based methods have gained significant attention in DMS. While~\cite{Schwung2022,Yuwono2023a} propose model-free and model-based state-based potential games, the works~\cite{Yuwono2024a,Yuwono2025} extend the approaches to leader-follower games. However, all these approaches rely on rather simple learning strategies without considering the potential of inverse models.

\subsection{Multi-Agent and Model-based Reinforcement Learning}\label{sec:rel3}

As we consider modular systems controlled by distributed controllers, the learning algorithm also has to be chosen in a distributed setting. However, extending single-agent to MARL~\cite{Busoniu2008} is not straightforward due to different obstacles discussed in~\cite{Kapoor2018}. Specifically, individually learning agents result in a non-Markovian environment for all agents, rendering the Markov decision process (MDP) only partially observable. Early approaches use cooperative settings based on Q-learning, assuming that the actions of other agents improve the collective reward~\cite{Lauer2000}. In recent years, various versions of deep MARL with a centralized critic have been proposed~\cite{Lowe2017,Foerster2017,3600270.3602057,3455716.3455894} resulting in the centralized training, distributed execution paradigm, see~\cite{amato2024introductioncentralizedtrainingdecentralized} for an overview. However, this requires extensive communication between the agents. In~\cite{Sunehag2017,Omidshafiei2017}, all agents share a common reward function, which requires constant inter-agent communication between all agents. In contrast, we consider fully distributed learning which allows for plug-and-play of modules in changing environments.

While the previously discussed works use model-free RL, the MBRL approach provides a promising alternative to reduce real-world interactions. MBRL frameworks can be roughly distinguished into world models~\cite{Ha2018} and planning approaches~\cite{Pinneri2020SampleefficientCM}. While the latter derives control policies similar to model predictive control using Monte-Carlo optimization, MBRL based on world models simultaneously learns an environment model and an optimal policy using standard RL algorithms. However, none of these approaches combine forward and inverse models to incorporate both the world model and policy network. The closest work to ours can be found in \cite{DIPRASETYA2025102945}, where inverse kinematic models have been incorporated into the policy networks of MBRL. However, the approach is restricted to systems modeled by kinematics, which is unsuitable for general dynamic systems.

\section{Distributed MBRL with inverse models}\label{sec:DMBRL}

In this section, we present the novel approach for distributed MBRL with inverse models. To this end, we first state the problem and give a system overview, followed by the inverse model architectures and the incorporation into RL algorithms.

\subsection{Problem Statement}\label{sec:problem}

\begin{figure}[htb]
 \centering
 \includegraphics[width=\columnwidth,keepaspectratio]{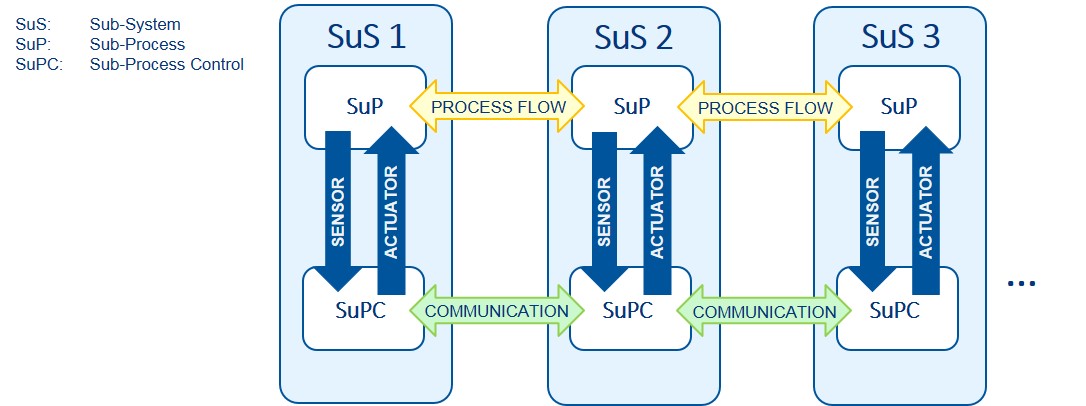}%[0 0 16 12] \includegraphics[width=\columnwidth,keepaspectratio]{Regelbasen.eps}
\caption{Considered system structure consisting of several subsystems with their own control system.}
\label{fig:sysstruct}
\end{figure}

We consider distributed systems consisting of several subsystems which can potentially be left out, included, or interchanged with other modules as illustrated in Fig.~\ref{fig:sysstruct}. Each module is equipped with its own local control system, which are able to communicate with neighboring modules. Note that such systems are common in modern manufacturing environments.

Based on the above system description, we state the following problem. Consider the distributed system $\mathcal{S}$ with $i=1,\ldots,n$ subsystems $\mathcal{S}_i$ equipped with local actuation $a_i$ as illustrated in Fig. 1. We define a number of production goals $m_i$ for each module indicated by $e_{i,j}$ with $i=1,\ldots,l$, $j = 1,\ldots, m_i$. Then, find the optimal production control over a given production episode $t = 0,\ldots,T$ by optimizing
\begin{align}
    e^*(t) = \max_{a_i} \sum^T_{t=0} \sum_{i,j} e_{i,j}(t).
\end{align}
The production goals $e_{i,j}$ can be arbitrarily defined, with typical examples including product concentrations in chemical reactors, mass flows in bulk good plants or processing times in manufacturing systems.

\subsection{Framework Overview}\label{sec:framework}

\begin{figure}[tb]
 \centering
 \includegraphics[width=\columnwidth,keepaspectratio]{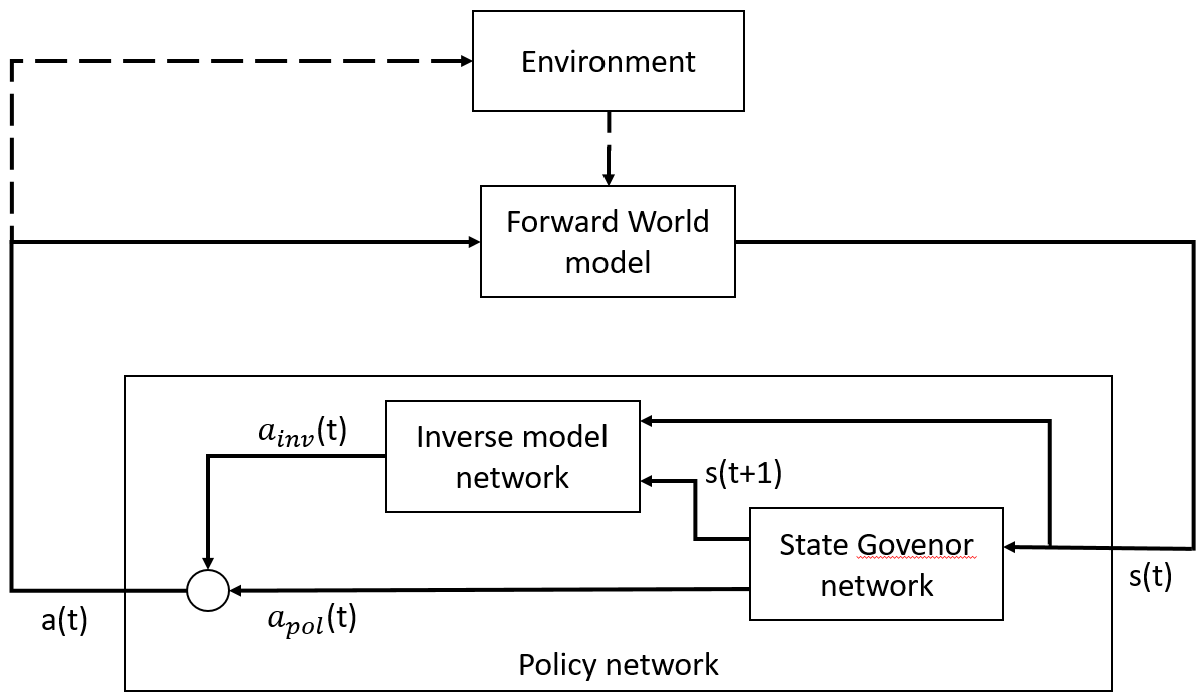}%[0 0 16 12] \includegraphics[width=\columnwidth,keepaspectratio]{Regelbasen.eps}
\caption{System overview of the proposed distributed model-based RL with inverse models.}
\label{fig:framework}
\end{figure}

After stating the problem, we now present our novel framework for distributed model-based RL, which is illustrated in Fig.~\ref{fig:framework}. First, we set up individual learning agents for each module or actuated system, resulting in $n$ independent agents that only share local state information with their neighboring agents, i.e., agent $i$ receives $s_{i-1},s_{i+1}$ in addition to its own state information only. The learning agents are implemented within an MBRL architecture using the world model approach~\cite{Ha2018}. In this setting, each agent first collects data by injecting sufficiently excited input signals and collects data sets $\mathcal{D}_i=\{s_i(k),s_{i-1}(k),s_{i+1}(k),a_i(k),s_i(k+1)\}_{k=0}^N$ consisting of states $s(k)$, actions $a_i(k)$ and state in the next time step $s_i(k+1)$. Using $\mathcal{D}_i$, we train a neural network model $f_{\theta_i}$ with parameters $\theta_i$ for the forward dynamics of the environment, i.e. we train $s_i(k+1)=f_{\theta_i}(s_i(k),s_{i-1}(k),s_{i+1}(k),a_i(k))$ using standard MSE-loss. In vanilla MBRL, after the forward model reaches sufficient accuracy, the policy network is trained on the forward model instead of training with the real environment, thereby strongly reducing real-world interactions.

The core novelty of the approach lies in the design and the training of the policy network. As illustrated in Fig.~\ref{fig:framework}, the policy network is split into two independent networks, namely an inverse model network and a state governor network. The inverse model network $g_{\gamma_i}$ with parameters $\gamma_i$ outputs actions $a_{inv,i}$ based on actual and next states, i.e. $a_{inv,i}(k)=g_{\gamma_i}(s_i(k+1),s_i(k),s_{i-1}(k),s_{i+1}(k))$. Note that we use the same data set $\mathcal{D}_i$ for the inverse model learning. Further, the inverse model is trained during the ramp-up phase together with the world model and is subsequently frozen during the MBRL learning stage. The state governor network corresponds to the policy network in the original MBRL setting, but with a different purpose. While in vanilla MBRL, the policy network learns to derive optimal actions to the environment, the state governor learns to derive the optimal next state to be visited by the agent. This optimal next state is subsequently sent to the inverse network to derive the required action to reach this state. Hence, this setting disentangles the learning of the task typically defined in the state space from deriving actions required to follow the state trajectories. In parallel, the state governor network also outputs a corrective action component $a_{pol,i}$ which is added to $a_{inv,i}$ to obtain the final action $a_{i}$.

\subsection{World models and Inverse world models}\label{sec:invmodel}

After discussing the general framework, we present details about the design and training of the world models. In general, the used network types can be chosen arbitrarily depending on the complexity of the system and inverse dynamics, i.e., we can potentially use standard MLPs, recurrent networks such as LSTMs, or autoregressive architectures like transformers. 

For the considered distributed MBRL set-up, where the individual dynamics of the subsystems are typically less complex, we found MLPs for both the forward world model and the inverse model to be sufficient to obtain both training stability and sufficient accuracy. Specifically, we define small decentralized networks, one for each actuator. Each network has a $4 \rightarrow 128 \rightarrow 128 \rightarrow 1$ architecture with SiLU activations and a linear output, which is clamped to the range $[0, 1]$. Both models are trained using MSE regression, i.e., we train
\begin{align}
    \theta_i^* \!\!=\! \min_{\theta_i} \text{E}_{\mathcal{D}_i}(\|s_i(k\!+\!1)\!-\!f_{\theta_i}\!(s_i(k),s_{i-1}(k),s_{i+1}(k),a_i(k))\|^2)
\end{align}
for the world model network and 
\begin{align}
    \gamma_i^* \!\!=\! \min_{\gamma_i} \text{E}_{\mathcal{D}_i}(\|a_{i}(k)\!-\!g_{\gamma_i}\!(s_i(k\!+\!1),s_i(k),s_{i-1}(k),s_{i+1}(k))\|^2)
\end{align}
for the inverse model. We remark that $\mathcal{D}_i$ only consists of one-step predictions. An extension to multi-step predictions is straightforward using e.g. sequence-to-sequence network architectures and will be considered in future work. We also remark that the training of both networks is done during the ramp-up phase only. Thereafter, both network parameters are kept fixed. We note that we experimented with keeping the inverse model parameters trainable, but with lower performance.

\subsection{Model Based RL with inverse models}\label{sec:RL}

After the model training, we now discuss the training of the policy network using distributed MBRL. To this end, first note that the core difference from the vanilla MBRL algorithm lies in the specific definition of the policy networks. In fact, the novel proposed policy networks essentially consist of the inverse model network $g_{\gamma_i}$ and the state governor network $h_{\theta_i}(s_i(k),s_{i-1}(k),s_{i+1}(k))$. Let $h^s_{\theta_i}$ and $h^a_{\theta_i}$ denote the two subnetworks with outputs $s_i(k+1)$ and $a_{pol,i}(k)$ respectively, the resulting policy network $\pi_{\theta_i}$ is given by (see Fig.~\ref{fig:framework}):
\begin{align}
    \pi_{\theta_i}(s) = h^a_{\theta_i} + g_{\gamma_i}(h^s_{\theta_i})
\end{align}
Note that in the above, we assume the inverse model network to be fixed, such that $\pi_{\theta_i}$ only depends on the learnable parameters $\theta_i$. 

Applying the specifically defined policy networks, we can straightforwardly employ arbitrary RL algorithms, online and offline, as well as value function-based, actor-critic approaches or the various proposed subversions to the above policy network training~\cite{9569056}. In the results, we will present a comparison of TD3~\cite{Fujimoto2018AddressingFA}, SAC~\cite{conf/icml/HaarnojaZAL18}, and DDPG~\cite{Lillicrap2016} for the considered system application. We focus on off-policy actor-critic methods, where the action-space pathologies the inverse model addresses are most pronounced.

Finally, Algorithm~\ref{alg:mbgt} provides the pseudocode of the implementation and learning procedure of MBRL with inverse models. 
\begin{algorithm}
\caption{Basic Approach of MBRL with inverse models.}\label{alg:mbgt}
\SetKwInput{kwEnv}{Environment}
\kwEnv{Real environment, World model}
\KwData{Actual episode for world model learning and MBRL, max. episode for world model learning and max. episode for MBRL, max. time duration per episode}
\KwResult{Trained $f_{\theta_i}$, $g_{\gamma_i}$, $\forall i=1,\ldots,N$,  selected action(s)}
\tcc{Model-based learning to derive world and inverse model}
initialize hyperparameters, $f_{\theta_i}$, $g_{\gamma_i}$, $\forall i=1,\ldots,N$\;
\While{max. episode has not been reached}{
    calculate initial states\;
    begin an episode\;
    \While{max. time duration has not reached}{
        obtain actual states from $f_{\theta_i}$\;
        calculate subsequent states and rewards\;
        generate new action(s) from $A_{gen}$\;
        store experience $(s_i(k),s_{i-1}(k),s_{i+1}(k),a_i(k),s_i(k+1))$ in buffer $\mathcal{D}_i$\;
        \If{buffer $\mathcal{D}_i$ is full}{
            sample random minibatch of experiences from buffer \textit{H}\;
            train $f_{\theta_i}$, $g_{\gamma_i}$, $\forall i=1,\ldots,N$, in a supervised manner\;
            }
        }
    }
\tcc{Training MBRL agent in world model environment}
Run an arbitrary RL algorithm to train state governor networks by replacing the real environment with world models $f_{\theta_i}$, $\forall i=1,\ldots,N$.
\end{algorithm}

\section{Experiments and Results}\label{sec:results}

We apply our methodology to a Bulk Good Laboratory Plant (BGLP). This section contains an overview of the BGLP, the training setup, experimental findings, and comparisons.

\subsection{Testing Environment: The Bulk Good Laboratory Plant}
\label{sec:bglp}

\begin{figure}[t]
	\centering
	\includegraphics[width=1.0\linewidth,keepaspectratio]{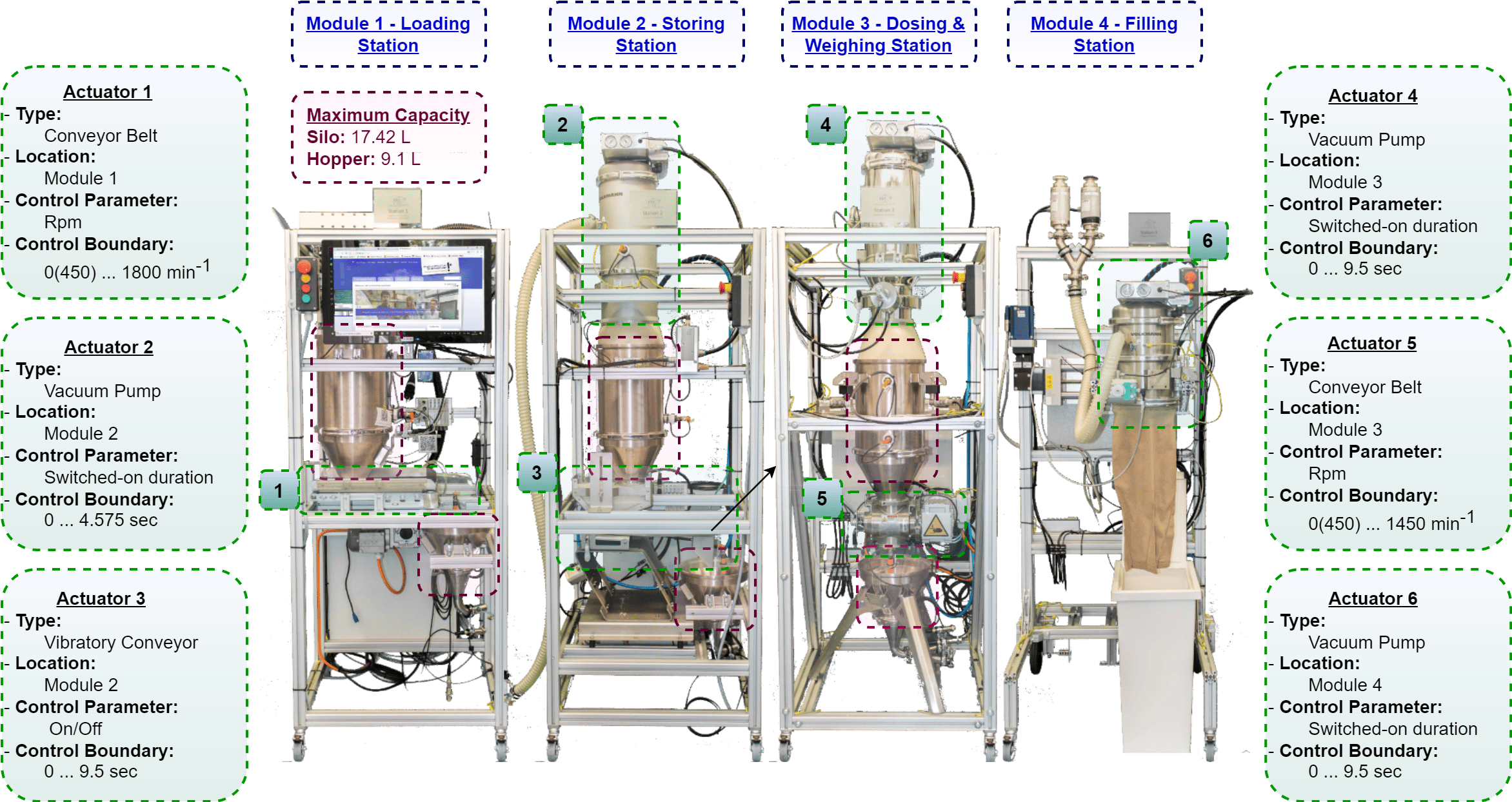}
	\caption{The Bulk Good Laboratory Plant, including the details of its modules, actuators, and reservoirs.}
	\label{fig:bglp}
\end{figure}

The BGLP~\cite{Schwung2022} is a distributed manufacturing system for bulk good transport through a network of various actuators and reservoirs. The system has four operational modules: loading, storage, weighing, and filling, as depicted in Fig.~\ref{fig:bglp}.

In Module 1, the control parameter for the conveyor belt is the motor rotation speed in rpm, similar to the rotary feeder in Module 3. The vibratory conveyor in Module 2 uses a binary value to indicate whether it is fully operational. For the vacuum pumps in Modules 2, 3, and 4, the control parameter is the turn-on duration in seconds. The simulation of the BGLP is available in the MLPro framework~\cite{Arend2022a}. The simulation is useful for training the proposed MBRL framework before deployment on the real system.

\subsection{Training Setup on the BGLP}
\label{sec:training}

Within the BGLP, each player $i$ corresponds to an individual actuator and has two pieces of state information, which are the fill levels of the preceding and subsequent reservoirs. Further, two objectives are defined: First, maintaining the fill level of the preceding and subsequent reservoirs to prevent bottlenecks and overflow, $r_{V}^i$, and second, minimizing power consumption, $r_{P}^i$. Additionally, the final player is tasked with an additional objective, such as meeting production demand, $r_{D}^i$.

The objectives are evaluated at each time step $t$, resulting in reward functions:
\begin{equation} \label{Eq:obj_1}
\begin{split}
    r_{V}^i = \frac{1}{1\!+\!V_{p}^{i}} + \frac{1}{1\!+\!V_{s}^{i}}; \quad
    r_{P}^i = \frac{1}{1\!+\!P^{i}}; \quad
    r_{D}^i = \frac{1}{1\!-\!V_{D}}; \\
\end{split}
\end{equation}
where $V_{D}$ is production demand fulfilment, $P^i$ power consumption, and $V_{p}^{i}, V_{s}^{i}$ local constraints to prevent bottleneck and overflow in the buffers. The production demand fulfilment $V_{D}$ is calculated as
\begin{align}\label{Equation:4-2-1_DemandVolume}
V_{D}=\smashoperator{\int_{0}^{T_{I}}}\dot{D}_t\,dt, \quad \dot{D}_t=
\begin{dcases}
\dot{V}_{N,out}-\dot{V}_{N,in}, &\text{if}\ h^{N}=0,\\
0, &\text{otherwise,}\\
\end{dcases}
\end{align} 
where $\dot{V}_{N,out}$, $\dot{V}_{N,in}$ denote the outflow and inflow to the buffer, $h^{N}$ is the normalized value of the fill level, and $T_I$ represents the duration of an iteration. The demand is considered fulfilled if the fill level of the last hopper exceeds the specified production demand. Meanwhile, the mathematical expressions for $V_{p}^{i}$ and $V_{s}^{i}$ are provided as follows:
\begin{equation} \label{Equation:4-2-1_Badewanne}
V_{p}^{i}=\smashoperator{\int_{0}^{T_{I}}}\mathbbm{1}_{q^i_{p}<Q^i_{p}}(q^i_{p})\,dt, \quad V_{s}^{i}=\smashoperator{\int_{0}^{T_{I}}}\mathbbm{1}_{q^i_{s}>Q^i_{s}}(q^i_{s})\,dt, 
\end{equation}
where $Q^i_{s}$ and $Q^i_{p}$ represent the upper and lower limits of the associated fill levels. 

Based on the above derivations, we apply reward functions adopted from~\cite{Schwung2022}, which are outlined as follows:
\begin{equation} \label{Equation:UtilityFunction}
 R_t^i=
    \begin{cases}
    \omega_v \cdot r_{V}^i+\omega_p \cdot r_{P}^i+\omega_d \cdot r_{D}^i & \text{if } i=N,\\
    \omega_v \cdot r_{V}^i+\omega_p \cdot r_{P}^i & \text{otherwise,}
    \end{cases}
\end{equation}
where $\omega_v, \omega_p, \omega_d$ are pre-defined weights of each objective. In this paper, we apply $\omega_v= 1.0, \omega_p= 0.001, \omega_d= 4.0$.

\subsection{Experimental Results}
\label{sec:res}

We now present the experimental results on both the inverse model training as well as the MBRL approach with inverse models. Every experiment is conducted under identical settings and scenarios within the BGLP, where we conduct 100 training episodes, with each episode spanning 10,000 seconds of production time (1,000 cycles) and targeting a production rate of 0.125 and 0.15 L/s. Then, we conduct an additional testing episode utilizing the reward function outlined in Eq.~\eqref{Equation:UtilityFunction}.

%Furthermore, each experiment undergoes hyperparameter tuning utilizing Hyperopt~\cite{Bergstra2012} with a random grid search algorithm. 

\subsubsection{Inverse model training}
\label{sec:reswim}

\begin{figure}[htb]
 \centering
 \includegraphics[width=0.8\columnwidth,keepaspectratio]{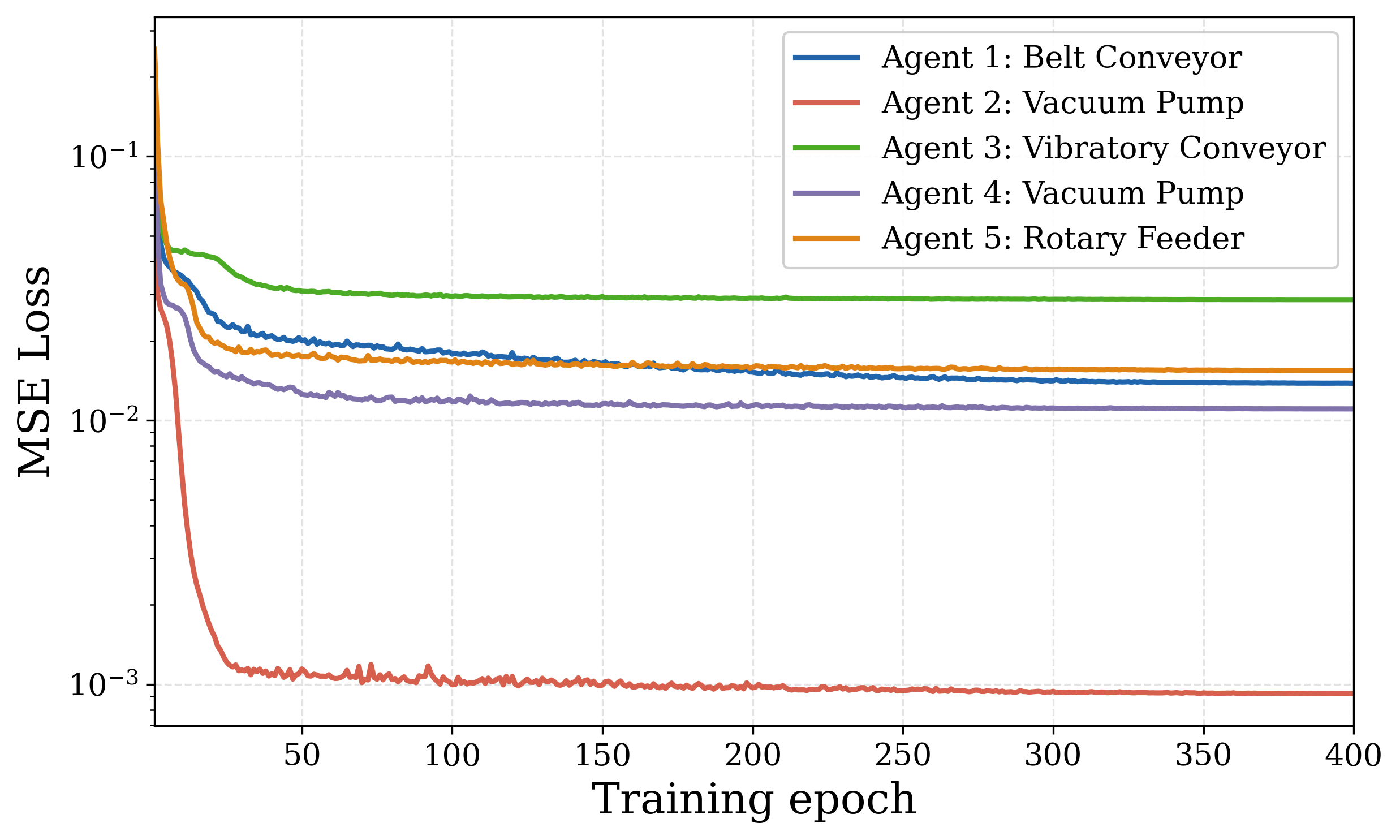}%[0 0 16 12] \includegraphics[width=\columnwidth,keepaspectratio]{Regelbasen.eps}
\caption{Training loss of the inverse models for each agent.}
\label{fig:trainingloss}
\end{figure}

\begin{figure}[htb]
 \centering
 \includegraphics[width=\columnwidth,keepaspectratio]{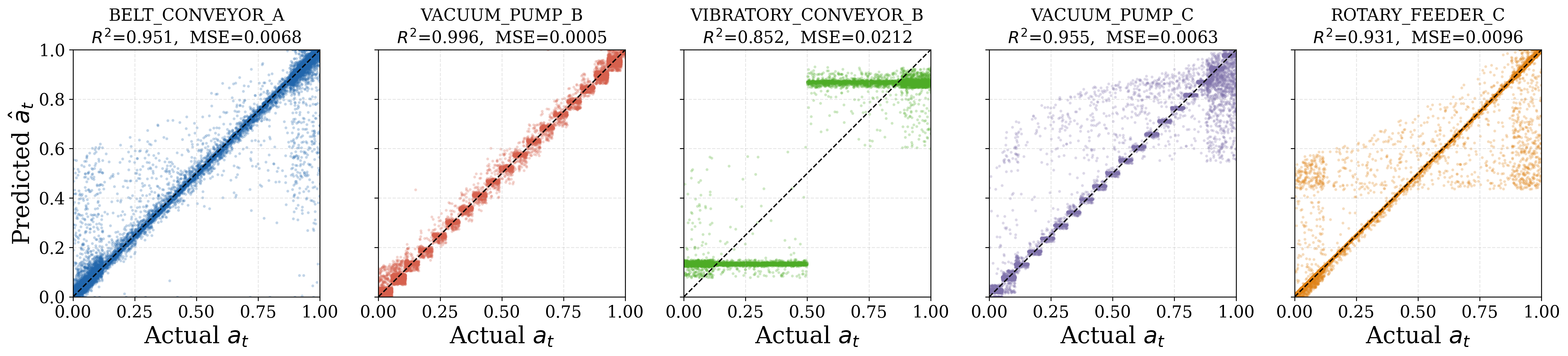}%[0 0 16 12] \includegraphics[width=\columnwidth,keepaspectratio]{Regelbasen.eps}
\caption{Comparison of inverse model outputs and target value for each actuator.}
\label{fig:scatterplayer}
\end{figure}

We start by reporting the results of the inverse model training during the ramp-up phase. Note that training results of typical world model training have already been reported in~\cite{Yuwono2023a} and are omitted due to space restrictions. The trajectories of the training losses for each actuator are presented in Fig.~\ref{fig:trainingloss}. The loss curves show a consistent decrease, indicating stable training with sufficient accuracy. Fig.~\ref{fig:scatterplayer} shows a comparison of predicted action and ground truth action using scatter plots. As can be seen, all actuators show a very good agreement with their ground truth value (along the straight line) also on unseen data. Note that actuator 3 is discrete operating, resulting in a step function reference.

\subsubsection{Distributed MBRL with inverse models}
\label{sec:reswim}

\begin{figure}[htb]
 \centering
 \begin{subfigure}[b]{\columnwidth}
  \centering
  \includegraphics[width=\linewidth,keepaspectratio]{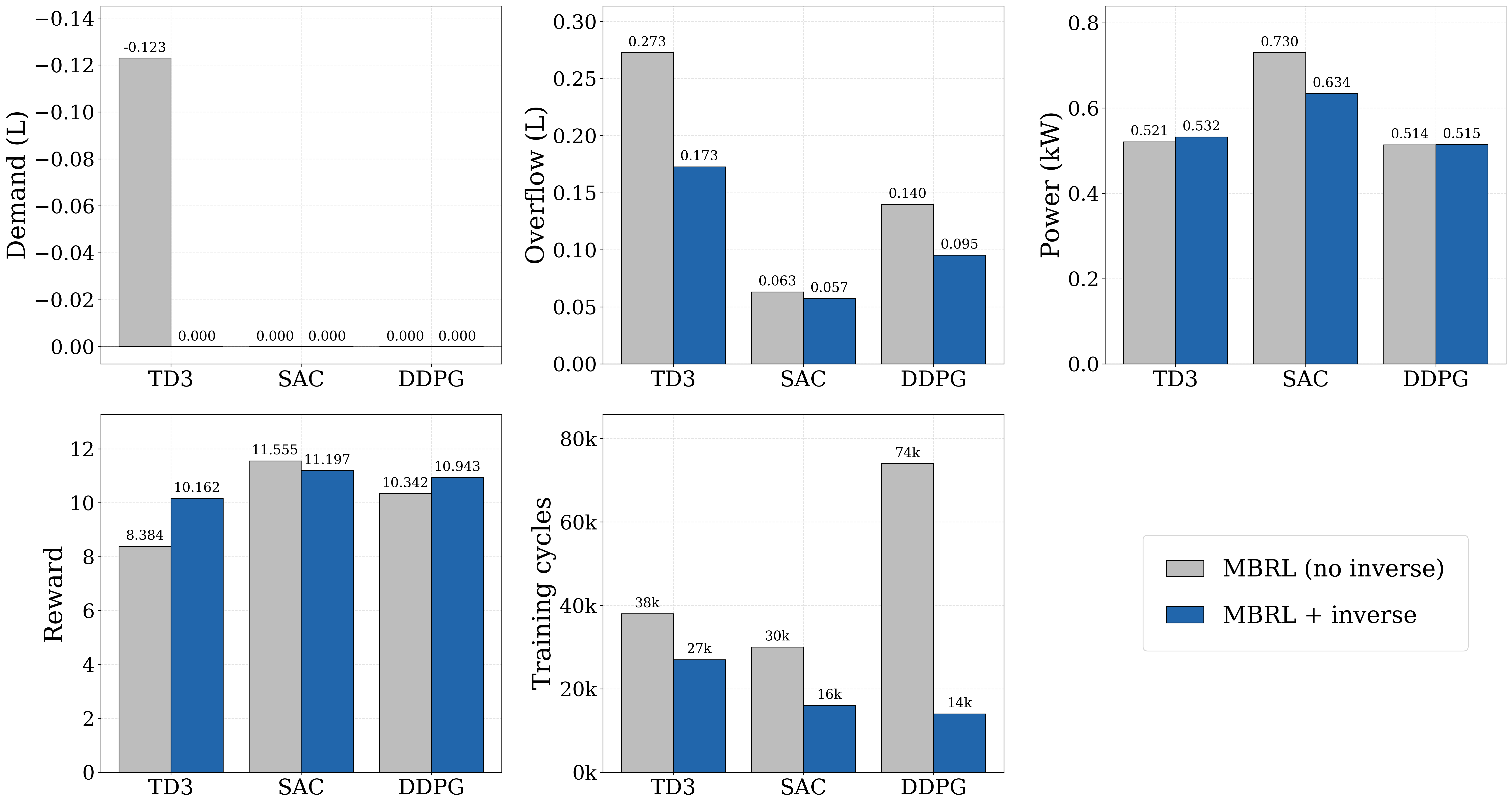}
  \caption{Production demand = 0.125~L/s.}
  \label{fig:MBRLcomp_a}
 \end{subfigure}
 \vspace{0.5em}
 \begin{subfigure}[b]{\columnwidth}
  \centering
  \includegraphics[width=\linewidth,keepaspectratio]{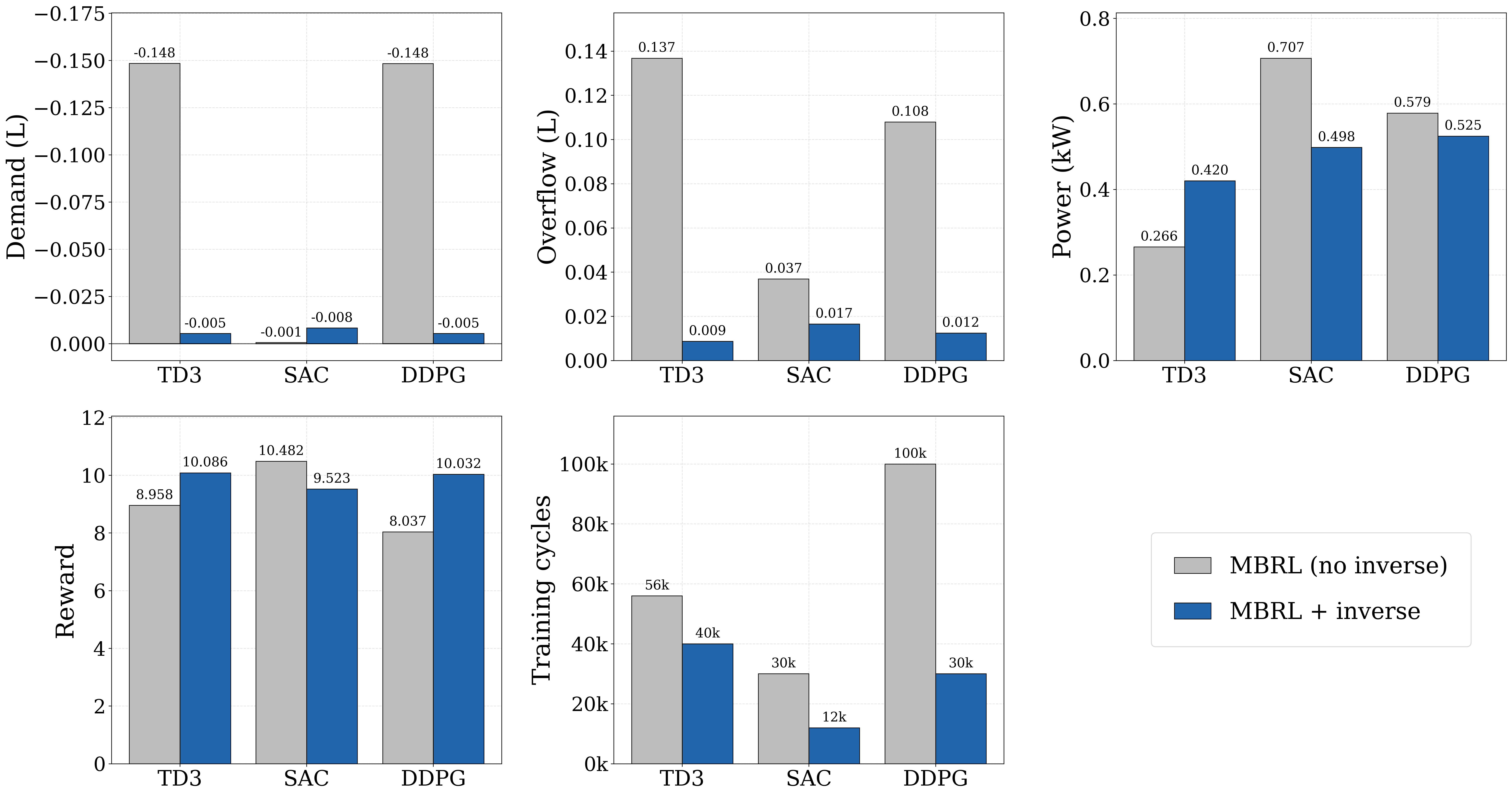}
  \caption{Production demand = 0.15~L/s.}
  \label{fig:MBRLcomp_b}
 \end{subfigure}
 \caption{Comparison of distributed MBRL approaches with and without inverse models on the BGLP for TD3, SAC, and DDPG.}
 \label{fig:MBRLcomp}
\end{figure}

Fig.~\ref{fig:MBRLcomp} compares the distributed MBRL approaches with and without inverse models for TD3, SAC, and DDPG at production rates of 0.125 L/s and 0.15 L/s. The reference "without inverse models" corresponds to the same MBRL setting in which the policy outputs actuator commands directly, i.e., the state governor and the inverse model are replaced by a standard policy network. We report demand fulfilment, overflow, and power consumption together with the overall reward and the required training cycles. Two effects are consistent across all six settings: the inverse models reduce the overflow (e.g., from 0.137 L to 0.009 L for TD3 at 0.15 L/s) and they reduce the required training cycles (e.g. from 100k to 30k for DDPG at 0.15 L/s). The effect on the overall reward is algorithm-dependent.

TD3 and DDPG do not solve the task without inverse models: they fail to meet the production demand and overflow heavily, which also renders their apparently low power consumption meaningless, since hardly any material is transported. With inverse models, both meet the demand and improve the reward from 8.0-9.0 to approximately 10.1. SAC, in contrast, already solves the task without inverse models. Here the reward is comparable, while overflow and power consumption are clearly reduced. As expected, the inverse model acts as a form of pretraining of the policy, which makes the overall training easier and results in shorter training times. We attribute the pronounced benefit for TD3 and DDPG to off-policy learning, such as bootstrapping from a replay buffer of raw actions, which makes these deterministic policy-gradient methods latch onto unstable action combinations, whereas learning desired next states delegates action generation to the frozen inverse model.

Regarding computational overhead and sensitivity, the inverse models add negligible cost: each is a small decentralized network ($4 \rightarrow 128 \rightarrow 128 \rightarrow 1$, roughly 17k parameters per actuator) trained only once during the ramp-up phase and kept frozen thereafter, so at policy-training and deployment time it contributes only a single additional forward pass per step. Regarding sensitivity to model inaccuracies, because the policy operates in task space and the inverse model only decodes a desired next state into an action, a bounded inverse-model error acts as a bounded actuation perturbation that the policy compensates for during training; the clamping of the output to $[0,1]$ additionally limits its effect. This is consistent with our observation that freezing the inverse model outperforms keeping it trainable, indicating robustness to moderate inverse inaccuracies.

%Finally, we report a comparison of different distributed MBRL algorithms in Fig.~\ref{fig:MBRLcomp}. We report individual rewards including demand fulfillment, overflows, and power consumption together with the overall reward and the required training cycles for the corresponding algorithms. As can be seen, MBRL with inverse models slightly outperforms their counterparts without inverse models. More importantly, the required training cycles until the results are reached are significantly lower for training with than without inverse models. As expected, the inverse model allows a form of pretraining of the policy, making the overall training easier, resulting in shorter training times.

\section{Conclusion and Future Work}\label{sec:concl}

This paper introduced a novel data-driven self-learning control framework for modular manufacturing systems based on MBRL. By incorporating inverse process models into the policy learning process, the proposed approach separates the learning of actuation dynamics from the dynamics in the task space, enabling RL to focus on task-space optimization, leading to more efficient policy training. The proposed framework was evaluated on a laboratory-scale testbed. The experimental results demonstrate that the integration of inverse models improves both learning efficiency and performance compared to conventional RL approaches, particularly for off-policy methods. Future work will focus on investigating alternative network architectures, specifically sequence-to-sequence architectures for inverse models, and analyzing the robustness of the approach under changing operating conditions and model uncertainties.

Beyond this, we plan to evaluate scalability on larger line-shaped modular plants and, ultimately, on the real testbed, and to complement the reported results with a statistical evaluation across multiple random seeds to establish confidence intervals on the observed improvements.

\bibliographystyle{IEEEtran}
\bibliography{references}

\end{document}